\documentclass[sn-mathphys]{sn-jnl}

\usepackage{graphicx}
 \usepackage{booktabs}

\usepackage{subcaption}
\usepackage{xcolor}
\definecolor{alizarin}{rgb}{0.82, 0.1, 0.26}
\usepackage{makecell}

\usepackage{hyperref}

\jyear{2021}%

\theoremstyle{thmstyleone}%
\theoremstyle{thmstyletwo}%

\theoremstyle{thmstylethree}%

\begin{document}

\title[Self-recognition in conversational agents]{Self-recognition in conversational agents}


\author*[1]{\fnm{Yigit} \sur{Oktar}}

\author[2]{\fnm{Erdem} \sur{Okur}}

\author[3]{\fnm{Mehmet} \sur{Turkan}}

\affil*[1]{ \city{Izmir}, \country{Turkey}}

\affil[2]{\orgdiv{Department of Software Engineering}, \orgname{Izmir University of Economics}, \orgaddress{ \city{Izmir}, \country{Turkey}}}

\affil[3]{\orgdiv{Department of Electrical and Electronics Engineering}, \orgname{Izmir University of Economics}, \orgaddress{\city{Izmir}, \country{Turkey}}}


\abstract{In a standard Turing test, a machine has to prove its humanness to the judges. By successfully imitating a thinking entity such as a human, this machine then proves that it can also think. Some objections claim that Turing test is not a tool to demonstrate the existence of general intelligence or thinking activity. A compelling alternative is the Lovelace test, in which the agent must originate a product that the agent's creator cannot explain. Therefore, the agent must be the owner of an original product. However, for this to happen the agent must exhibit the idea of self and distinguish oneself from others. Sustaining the idea of self within the Turing test is still possible if the judge decides to act as a textual mirror. Self-recognition tests applied on animals through mirrors appear to be viable tools to demonstrate the existence of a type of general intelligence. Methodology here constructs a textual version of the mirror test by placing the agent as the one and only judge to figure out whether the contacted one is an other, a mimicker, or oneself in an unsupervised manner. This textual version of the mirror test is objective, self-contained, and devoid of humanness. Any agent passing this textual mirror test should have or can acquire a thought mechanism that can be referred to as the inner-voice, answering the original and long lasting question of Turing ``Can machines think?'' in a constructive manner still within the bounds of the Turing test. Moreover, it is possible that a successful self-recognition might pave way to stronger notions of self-awareness in artificial beings. }




\maketitle

\section{Introduction}
In $1950$, Alan Turing investigated the question, ``Can machines think?''. However, instead of providing a concrete definition of ``thinking'', he replaced the question with a test called the Imitation Game (IG) claiming that a machine passing this test can then be regarded as a thinking entity~\cite{tur1950}.

\begin{figure*}
    \centering
    \begin{subfigure}[b]{0.32\textwidth}
            \includegraphics[width=\textwidth]{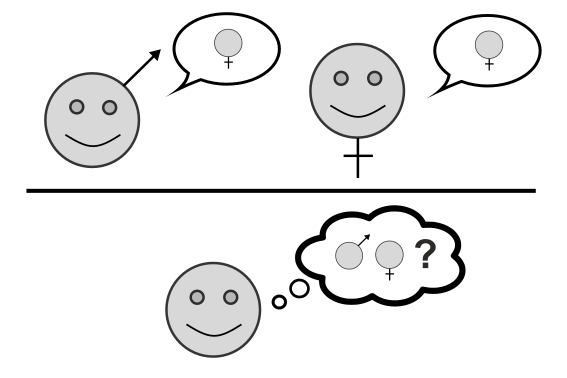}
            \caption{Preliminary of the Imitation Game}
            \label{fig:a}
    \end{subfigure}
    \begin{subfigure}[b]{0.32\textwidth}
            \includegraphics[width=\textwidth]{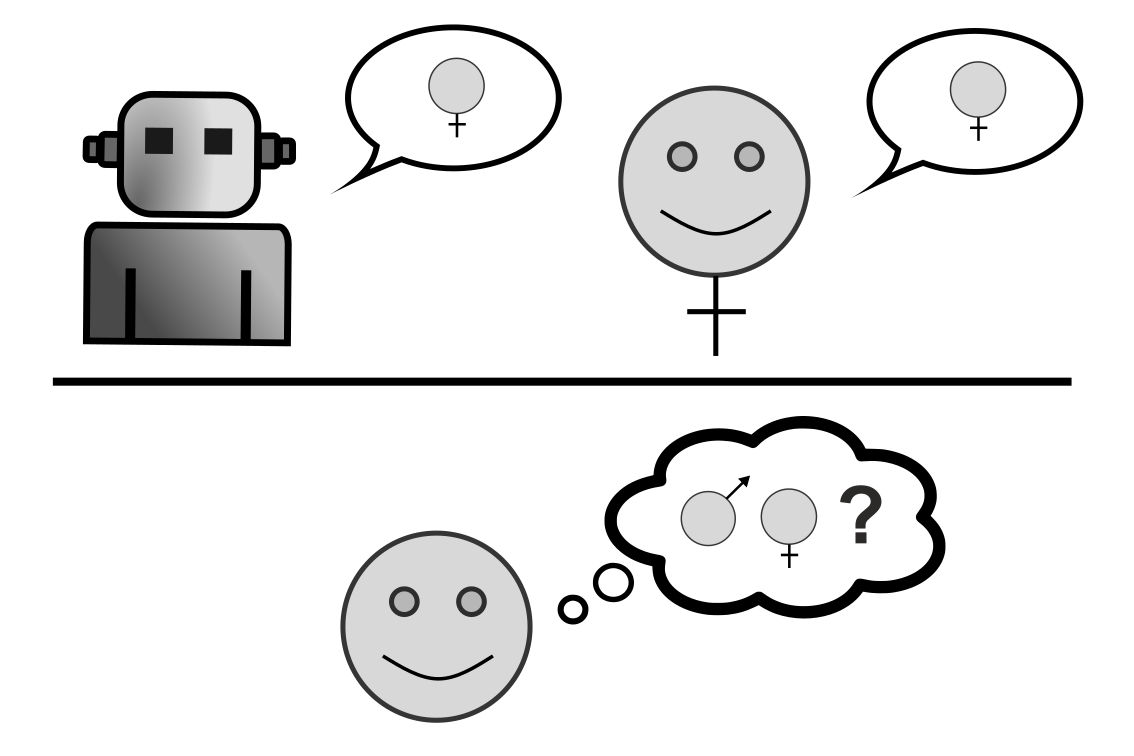}
            \caption{First version of the Imitation Game}
            \label{fig:b}
    \end{subfigure}

    \begin{subfigure}[b]{0.32\textwidth}
            \includegraphics[width=\textwidth]{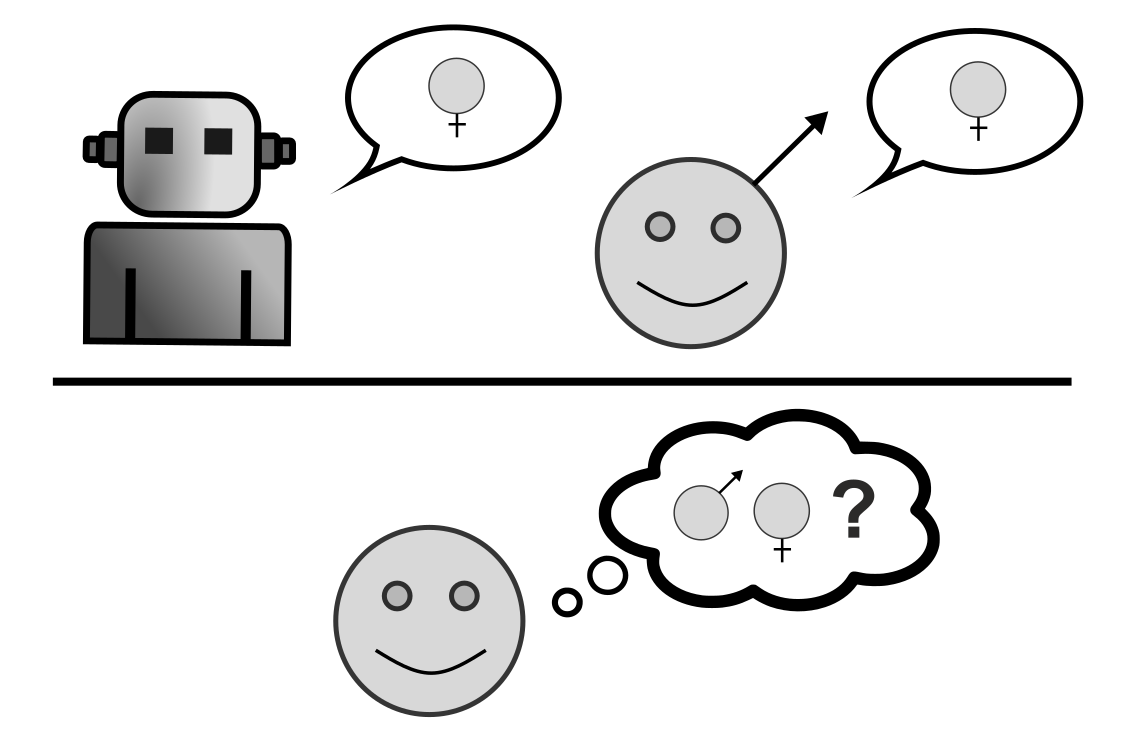}
            \caption{Second version of the Imitation Game}
            \label{fig:c}
    \end{subfigure}
    \begin{subfigure}[b]{0.32\textwidth}
            \includegraphics[width=\textwidth]{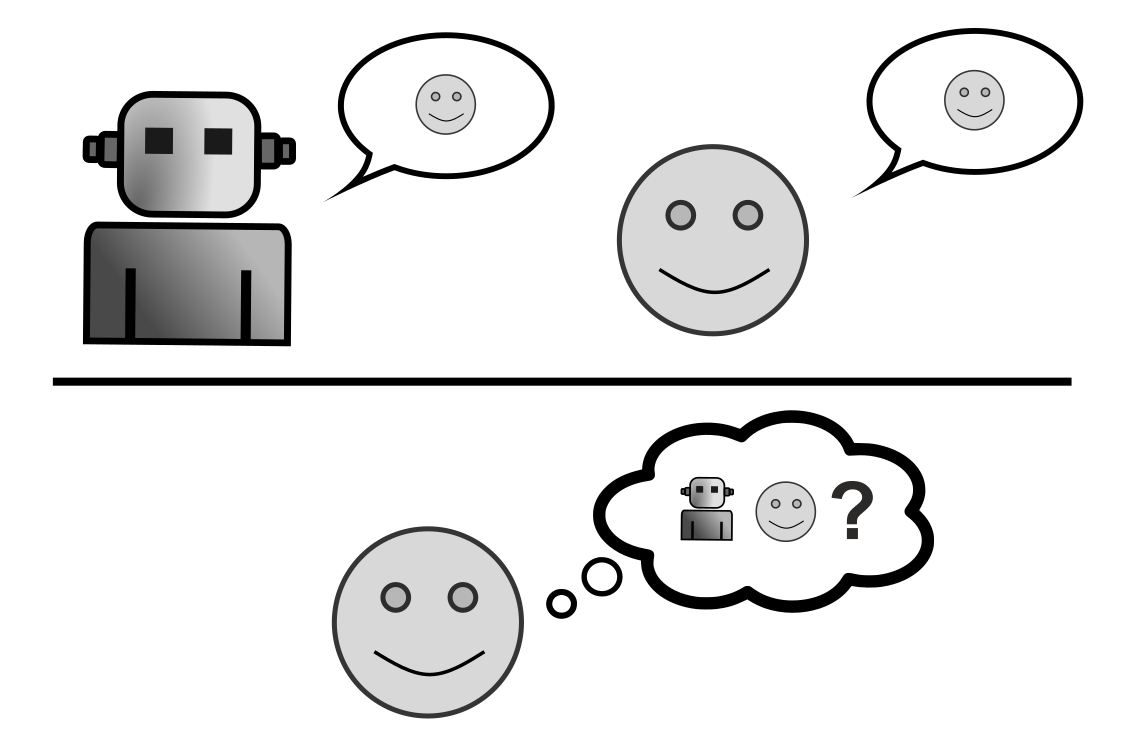}
            \caption{Standard version: The Turing Test}
            \label{fig:d}
    \end{subfigure}
    \caption{The distinction between the Imitation Game and the Turing Test}\label{fig:ABCD}
\end{figure*}

Originally being a party game, the IG is played with a man, a woman and a judge whose gender is not important as depicted in Fig.~\ref{fig:a}. Through written communication, both subjects aim to convince the judge that he/she is the woman and the other is not, hence the man needs to imitate a woman. Turing then replaces the man with a machine as in Fig.~\ref{fig:b} and wonders whether the success rate would change or not. In the final version of the game, a man now takes the place of the woman as in Fig.~\ref{fig:c}. 

It is not very clear which version (Fig.~\ref{fig:b} or Fig.~\ref{fig:c}) is meant by IG throughout the literature, but it is generally assumed that the Turing test (TT) in its standard interpretation takes the form of Fig.~\ref{fig:d} in which the machine's ability to imitate a human, instead of its ability to imitate a woman is measured. At first, it might be in question why IG Turing originally suggested is gender-based or whether IG is totally equivalent to TT. Readers might refer to~\cite{PinarSaygin2000} for a more comprehensible review of TT. Note that, Turing himself later drops gender-related issues and poses the question ``Can machines communicate in natural language in a manner indistinguishable from that of a human being?'', thus places TT as in Fig.~\ref{fig:d} into central attention~\cite{tur1950}. Consulting other Turing authored sources confirms that IG specific issues are not relevant for a general discussion of TT. 

\subsection{Analysis of Turing Test}
\label{sec:att}

In Table~\ref{tab:table1}, TT mostly refers to behavioral aspect of human intelligence but in fact targets the reasoning domain. In other words, TT claims that fulfilling Acting Humanly (AH) implies the existence of a thinking mechanism, namely Thinking Humanly (TH) or Thinking Rationally (TR), or both. This table is visited often throughout this subsection~\cite{Russell2009}. Note that, an analysis on TT in its standard form is given here, but the methodology presented later is more suited towards the viva voce version of the test. In such a version, human subject is discarded. We prefer to call such a version one-to-one as vica voce in dictionary terms means orally and may cause confusion. In that sense, throughout the text, standard TT and its one-to-one (i.e. one judge, one agent) version are considered mostly interchangeably. 

\subsubsection {Consistent machines, G{\"o}del and paraconsistent logic.}
As built on logic, conventional machines are bound by the G{\"o}del's Theorem which states that in consistent logical systems of enough power, certain statements cannot be proved or disproved within the system~\cite{godel1931formal}. Thus, they are bound to be limited to an extent. On the other hand, humans with some room for inconsistency or irrationality, have different characteristics. Implications of these issues on machine thought is further discussed by Lucas~\cite{lucas1996minds}. However, with introduction and mechanical realization of paraconsistent logic, which allows inconsistencies in a controlled and discriminating way, it is probable that such machines can then reason much like humans do~\cite{priest2002paraconsistent,costa2005}. Note that modeling generic human reasoning mechanism exactly in terms of paraconsistent logic is perhaps still an open problem. Then assuming it can be solved, referring to Table~\ref{tab:table1}, for artificial machines both TH and TR seem satisfiable. This conclusion is important with respect to TT as TT itself claims that TH and TR are satisfiable (i.e., by fulfilling AH).

\subsubsection {Machines that cheat, Chinese room and hard-coding}
In TT, there is no restriction on the design of the involved machines. Therefore, machines can cheat (i.e., hide their default behavior) and disguise themselves in a human-like appearance. In Table~\ref{tab:table1}, this corresponds to the fact that although AH does not hold in reality, the machine can trick the judge believing in that AH holds. Then by fulfilling AH, the machine can erroneously pass TT. A related argument is given by John Searle through his Chinese room example, claiming that external behavior cannot be used to determine whether a machine is actually thinking in real-time or reanimating an already thought and saved process (created by someone else)~\cite{searle_1980}. This translates to the fact that, although neither of TH, TR, AH or AR hold directly for the machine in question, it can simulate such processes (acquired from someone else) without actually having created them in the first place, working much like a virtual machine. Note that the capacity of such a machine is then bounded by the processes it has or can acquire, and thus its intelligence is not defined by its own mental abilities but through the abilities of its acquaintances. A chatbot named Eugene Goostman may be given as an example ~\cite{goostman}. He is accepted to  have passed the TT in a competition held by University of Reading to mark the 60th anniversary of Turing's death ~\cite{ttContest2014}. He made \%33 of 30 judges believe that it is human after five minutes of conversation. Furthermore, this success was published in several online tech news websites and TIME ~\cite{goostmanNewsReading,goostmanNewsZdnet,goostmanNewsTIME}. However, in reality the chatbot was only using its so called personality - a 13 year old Ukrainian boy - as a cover for its lack of real intelligence. It used humour and some personality quirks by stating its age to avoid some of the judge's questions. This to us is not a proof of showing real intelligence. It is just giving plausible answers to the questions and when the answer is not logical use personality and cleverly chosen age for that personality as an excuse. To show some real intelligence, a chatbot should use incoming query as an input with its knowledge prior on its own without such gimmicks discussed before to originate some answer that will satisfy the judge.

\begin{table}[t!]
  \begin{center}
    \caption{Summary of the extent of mainstream AI}
    \label{tab:table1}
    \begin{tabular}{l|l|l} 
        & \textbf{Human Intelligence} & \textbf{Rationality} \\
      \hline
      \textbf{Reasoning} & \makecell{Thinking \\Humanly (TH)}  & \makecell{Thinking \\Rationally (TR)}\\
      \textbf{Behavior} & \makecell{Acting \\Humanly (AH)} & \makecell{Acting \\Rationally (AR)}\\
    \end{tabular}
  \end{center}
\end{table}

Having roots in Lady Lovelace's objection, machines are generally assumed to be incapable of originating anything, doing anything new, or surprising us. Such acts most probably require learning and creativity besides keeping a set of logical rules. However, at current stage of AI research, machines (or artificial agents) are capable of learning and also undergoing tasks that require creativity~\cite{toivonen2015data,roberts2018magenta}. On the other extreme, Ned Block proposes a machine that can pass TT, without any significant information processing, but through being extensively hard-coded~\cite{Block1995}. This hypothetical machine stores all the possible sensible conversations in its memory, and then answers the judges just by simple lookups. Although such a machine may not be possible in practice, it is theoretically possible. This machine's intelligence is analogously equivalent to that of a jukebox, but it can pass TT, thus TT cannot be a proper test to measure intelligence. Block in fact claims the opposite of Turing, namely acting humanly does not necessarily imply the existence of any thinking activity, as in theory generic human behavior can be hard-coded into the machine. In short, TT is deemed to be a behaviorist approach not capable of detecting the extent of internal information processing. Therefore, alternatives that can detect the existence of sophisticated internal mechanisms (such as capability of learning and general problem solving) must be preferred. 

\subsubsection {Human intelligence vs. general intelligence.}
It is asserted that the IG (or TT) examines machines in terms of human-specific intelligence, instead of on the grounds of a general one~\cite{millar1973point}. This issue is investigated in detail when the concept of sub-cognition is introduced in Sec.~\ref{aap} and is also revisited when information theoretic alternatives of TT are discussed in Sec.~\ref{ita}. 

\subsubsection {Impairment of judges and confederate effect.}
Apart from the philosophical aspects of TT, involvement of judges is another issue from a practical perspective. As human beings, judges can make mistakes, cannot be totally objective and can even be manipulated towards an unexpected decision~\cite{tur1950,shah2010hidden,Shieber1994}. Moreover, human participants may frequently be demotivated to act as themselves, causing them to be incorrectly labeled as machines, such peculiarity being named as the confederate effect~\cite{warwick2015human}. In short, these all are repercussions of TT not being a self-contained test. There exists an external dependency on the performance of judges and also human subjects, thus TT cannot truly be accurate or objective. 

\subsection{Alternatives to Turing Test}
\label{att}
Alternatives to TT are investigated under three headings. Firstly, alternatives that provide valuable analytical insight are given. Then, higher-order generalizations of TT are discussed. More formal information theoretic alternatives are listed, then creativity is considered later leading the way to self-recognition as a substructure.

\subsubsection{Alternatives as analytic probes.}
\label{aap}
Considering the possibility of a random state finite automaton (FSA) to generate proper English sentences (by extreme luck) enough to pass TT, Kugel introduced a theoretical game consisting of infinitely many rounds~\cite{kugel1990time}. The main motivation behind is the fact that not only FSAs but even Turing machines should definitely be regarded inferior compared to mental capabilities of humans. Therefore, the possibility of a random FSA being able to pass TT in theory may be disturbing for some people~\cite{kugel1990time}. 

Through introduction of another hypothetical test, called the Seagull test, the limits of TT is further challenged. A Seagull test measures a subject's capability of flight. The subject will pass the test if its flying characteristic is indistinguishable from that of a seagull in the radar. The Seagull test cannot be fulfilled by helicopters, bats or beetles, or many other flying things alike. Thus, this is directly analogous to TT's detection of intelligence practiced by a human being. Further through the introduction of the sub-cognitive questions, French claims that to imitate a human, a machine needs to experience the world as a human does (i.e., through sensing organs) for a considerable period of time, otherwise it will not be able to answer questions related to special physical experience that can only be acquired by a human being. This is due to the fact that, while TT is a test for human-like intelligence and experience, just as the Seagull test is for detecting Seagull-like flight characteristics~\cite{french1990subcognition}. 

An objection to this claim is raised through an algorithm called PMI-IR (Pointwise Mutual information and Information Retrieval)~\cite{turney2001answering}. This ``disembodied'' algorithm has access to a search engine and can answer sub-cognitive questions using statistical analysis. However, it is in question whether going online or having access to a large collection of related text still counts as disembodiment. In other words, this algorithm is not a standalone product but depends totally on the large set of text as its source of knowledge. Namely, the boundaries of the algorithm is not clearly set. Although this seems as a solution to sub-cognition problem, it is harder to address further generalizations of sub-cognition concept as next.

\subsubsection{Generalizations.}
Based on the sub-cognition concept, it is natural to extend TT into the physical domain as proposed by Harnad~\cite{harnad1989minds} referred to as Total Turing test. In this physical version, judge can also directly, visually, tactically examine the two candidates. Harder extensions referred to as T4 and T5 are also discussed in~\cite{french2000turing}, but all these generalizations aim at testing human-specific intelligence or functionality of machines, not providing a universal measure for intelligence or capability. Therefore, information theoretic approaches to defining intelligence are discussed next to provide a domain independent discussion.

\subsubsection{Information theoretic alternatives.}
\label{ita}
A conventional perspective in this domain favors inductive learning capacity as a general and fundamental intelligence sign. Through certain analogies it can be claimed that inductive learning is tightly related to compression ability, thus it is possible to draw parallels between intelligence and algorithmic information theory~\cite{dowe1997computational}.

In a similar study, comprehension (as the outcome of a successful inductive inference process) is chosen as the fundamental sign of intelligence. Formalizing this ability, authors arrive at what is called the C-test, defined in pure computational terms applicable to both humans and non-humans. In this way, they are also able to establish a connection between information theoretic concepts and classical IQ tests~\cite{hernandez2000beyond}. 

Recent understanding suggests that a compression or induction test is possibly limited to define the standards of general intelligence. The key idea is to see intelligence as the mean (or weighted average) performance of an agent in all the possible environments~\cite{hernandez2010measuring}, including active environments. In this regard, the agent does not only require inductive abilities to understand the environment, but also needs planning abilities to use such understanding effectively. These modifications put focus on concepts such as perception, attention, and memory besides inductive skills, and thus generalizes C-test. Following this logic, a universal definition of intelligence is given as ``the ability to adapt to a wide range of environments'', both referring to internal and external mechanisms of an agent. Although theoretically sound, developing a practical test to satisfy such an extensive measure for all possible intelligence forms accurately and objectively is nearly impossible. 

\subsubsection{Creativity, and the Lovelace Test}

Lovelace test proposed (in the honor of Lady Lovelace) by Bringsjord et. al aims at measuring the creativity within the system~\cite{bringsjord2003creativity}. When the human creator of the system cannot account for the original product of the system then the system passes the test. This kind of logic is very relevant in investigation of current generative systems. Namely, generative adversarial networks (GAN)~\cite{goodfellow2014generative}, proposed as generative models for unsupervised learning, can generate new samples that look authentic. A more recent creative approach includes the third generation of generative pretrained transformers, namely GPT-3. Although GPT-3 is not targeted towards passing TT, it is surely capable of generating human-like text~\cite{floridi2020gpt}. This brings up the question of correctly determining who is the source of the generated product. Namely, in supervised machine learning, what matters is the expert labeling of training data. Although a machine can learn from such a labeling, the power lies within the expert that has performed the labeling beforehand. In generative models, that are based on unsupervised learning, there is no such background expert, and they are closer to passing the Lovelace test. We claim that, for a system to pass the Lovelace test, the system must be able to distinguish oneself from the others, and claim oneself as the originator of the product or the idea. Namely, one must exhibit the idea of self to distinguish oneself from the other beings, and most importantly distinguish oneself from one's creator (to be able to pass the Lovelace test). At this point, we then investigate whether sustaining the idea of self is possible within the TT or within its one-to-one version.

\subsection{One-to-one and inversion}
In that light, going back to TT and observing it as a practical test is inevitable. In fact, targeting one-to-one version of TT as the starting point of our methodology is the best we can do at this point. Without the adversarial input, the judge and the subject are isolated. Is it possible to go further to make such a system that depends on fewer interactions and parameters? Is it possible to design a digitally interesting and capable scenario that is independent of human interaction? 

Towards this aim, the inverted version of TT is one step closer to our scenario~\cite{watt1996naive}. In a more recent study, the subject is forced to pose questions to the judge, inverting the interrogation process~\cite{damassino2020questioning}. In this context, the concept of naive psychology comes to attention, which points to the natural tendency of ascribing mental states to other beings and themselves. This can be a faculty that offers an evolutionary advantage to animals in living complex societies~\cite{watt1996naive}. In an inverted TT, if a machine as a judge can judge the cases correctly, then passes the test. However, this poses a problem of identity when the machine is asked to discriminate between a human and an instance of oneself. Another objection to inverted TT is that it can be carried out successfully within a standard TT, where this is in parallel to the fact that universal Turing machine is a powerful formulation where it can simulate other equivalent systems~\cite{french1996inverted}. Such considerations bring us closer to the formulation we have in mind. Our methodology can be regarded as a one-to-one version of an inverted Turing test where the subject is oneself (either as a different instance, a mimicker, or as exactly oneself through a ``textual mirror''). It can be carried out in a standard Turing test, namely the judge can decide to copy and paste the responses back to the terminal as is, therefore act as a ``textual mirror''. This realization suggests that our test can be regarded as a subtool (i.e. a textual mirror) within TT, instead of being an alternative.

Up to now, we have investigated different dimensions of intelligence. To put it simply, we observe that there is an individualistic emphasis and yet there is also the social aspect of intelligence. In this sense, a mirror (in any dimensions) might be a practical tool to make the bridge between the individualistic versus the social aspect of intelligence. Therefore, in our formulation we regard self-recognition as a substructure (possibly optional) of the general intelligence that an entity can possess.

\section{Self-recognition as a substructure of intelligence}
\label{sraaa}
At this point, it is now most appropriate to introduce our perspective on this issue. Perhaps, instead of trying to attain a definition for the supreme form of intelligence and an ultimate test for it, trying to catch the glimpse of general intelligence will be far more beneficial as a cornerstone. It is a common convention to regard human beings as the most intelligent forms known to exist. Forcing machines to reach this (highest-known) intelligence threshold without setting any other reasonable lower milestones seems to be unreasonable. Assuming that species observed as the most intelligent ones of all have the ability of self-recognition in a mirror~\cite{reiss2001mirror,plotnik2006self,anderson2015mirror}, such tests and their variants can then be proper candidates for catching glimpses of general intelligence in a self-contained and a truly objective manner, immune to most of the objections raised against TT and its generalizations, setting a more reasonable milestone for machines to reach. 

\subsection{Self-recognition in living beings}
Self-recognition and idea of self is special and unique for some mammals like great apes, humans, orcas, dolphins and elephants. Also, there is also a non-mammal animal, ``European Magpies'', that is capable of self-recognition. First usage of ``the Mirror Test'' as a tool to test self-recognition appears in $1970$, introduced by Gordon Gallup~\cite{gallup1970chimpanzees}. Basically, a test subject is marked with an odorless dye, where it cannot see directly (forehead, ear etc.). Then, its behavior is observed to see whether it will be aware that the dye is on its own body part. For instance, one of the common behavior that indicates self-recognition is poking the marking on its own body. Of course, this behavior needs to happen while observing the reflection. However, the test raised some questions and received critics. Readers may refer to~\cite{sue2006self} for more detailed information on the upcoming discussion about self-recognition and interpreting others' behaviors. 

The test was not robust enough. Gorillas, for instance, failed on the test because their basic instincts dictate that eye contact is an aggressive gesture. Hence, they avoid looking at the reflection in the face. Other than this, some of the primates need a transitional period before self-recognizing themselves in the mirror (like systematically exploring the body parts that cannot be seen directly). These types of problems complicate the application procedure of the test. Nevertheless, the main question was ``Is Mirror Self-Recognition enough to say the subject is intelligent?''. An early attempt of an answer came from Gordon Gallup again. According to Gallup, there was a link between MSR and ability to interpret others' mental states. In short, he predicted a developmental correlation between presence of mirror self-recognition and social strategies based on the idea of self, such as empathy, pretending, and deception, to be reached with some cognitive development after presence of mirror self-recognition. To best of our knowledge, it can be thought as a door that leads to more complex cognitive abilities. It should also be noted that, mirror self-recognition presence is needed for more complex cognitive abilities but it is not a certainty that an animal with such ability will develop more complex cognitive abilities ever~\cite{gallup1998self}. 

Recent studies suggest that studying MSR in a gradual manner might be more realistic, instead of observing it as a binary case~\cite{de2019fish}. The reason for that is, there are studies which have engineered MSR into monkeys~\cite{chang2015mirror} and cleaner fish ~\cite{kohda2019if} through irritant markers. Both monkeys and cleaner fish cannot pass the original mark test. Such studies originate the concept of felt mark test, forcing self-recognition concept into a gradual domain. Moreover, in a more stringent study, monkeys are able to pass MSR after learning precise visual-proprioceptive association for mirror images, without the need for an irritant marker. It is claimed that learning to use a mirror as an instrument for mirror-induced self-directed behavior may help us understand the neural mechanisms of MSR~\cite{chang2017spontaneous}.

Perhaps the most promising way to investigate self-recognition issue, both in practical and theoretical terms, is through studying mirror neurons that can identify actions of either self or others~\cite{kohler2002hearing}. Namely, such audiovisual mirror neurons activate when the corresponding action is performed, its related sound is heard, or it is seen. A recent study successfully incorporates such findings and develops a brain-inspired model for robotic self-consciousness~\cite{zeng2018toward}. However, common conception is that mirror neurons alone are not sufficient, but only necessary for the ability of self-recognition in a mirror, as monkeys possess such neurons but cannot pass the mirror test in its conventional form. Let us investigate self-recognition problem from robotics perspective now, to be able to have more theoretical understanding.

\subsection{Robotic self-recognition}
\label{rsr}
In one of the earliest studies towards robotic self-recognition, through learning a characteristic time window between the initiation of motor movement and the perception of actual motion, authors demonstrate a certain level of self-recognition in robots, reminiscent of a rather incomplete model for self-awareness present in human infants~\cite{michel2004motion}. However, in case of co-occurence of very similar time delay characteristics for two different agents, such system will fail as there is then little chance of discrimination. After all, motion time delay models are certainly not as unique as fingerprints, namely two robots of the same kind from the same manufacturer will surely have very similar time delay models.

In a follow-up study from the same research group, the robot is now able to learn a more formal Bayesian model that relates its own motor activity with perceived motion. The importance of this study lies in the fact that mirror self-recognition is performed through a purely statistical kinesthetic-visual matching mechanism, without any significant social aspect. This is a challenge to the viewpoint that a certain level of social understanding is necessary for mirror self-recognition. Authors further claim that mirror test may not be about self-awareness or theory of mind at all, but merely a test of an agent's ability to adapt to new kinds of visual feedback, but such an ability might be related to intelligence, mind, self-awareness concepts in a self-referential way (recalling the rather synchronous definition of general intelligence as the ability to adapt to a wide range of environments as given earlier).

In this study, the center of attention is not visual self-recognition as exemplified up to now, but a conceptual version of it. A simple consideration suggests that mirror neurons can be used for implementing a communication system based on gestures, as in a sign language, thus introducing language into our research domain. There are many studies on the relationship between mirror neurons and language~\cite{fogassi2007mirror,stamenov2002mirror}. With all these considerations, it is time to present our proposed methodology which conceptualizes the mirror test by providing it in a textual/conceptual form.

\subsection{Proposed Methodology}
\label{sec:3}

\begin{figure*}[t!]
    \centering
    \begin{subfigure}[b]{0.4\textwidth}
            \includegraphics[width=\textwidth]{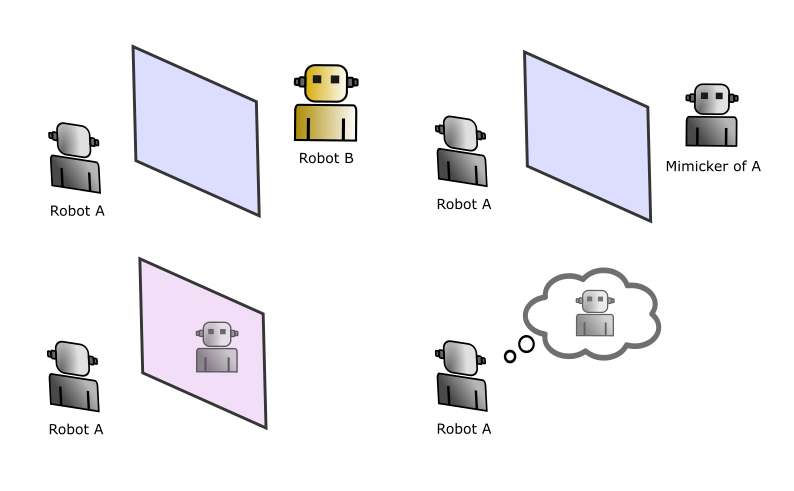}
            \caption{Visual version}
            \label{fig:vis}
    \end{subfigure}
    \begin{subfigure}[b]{0.4\textwidth}
            \includegraphics[width=\textwidth]{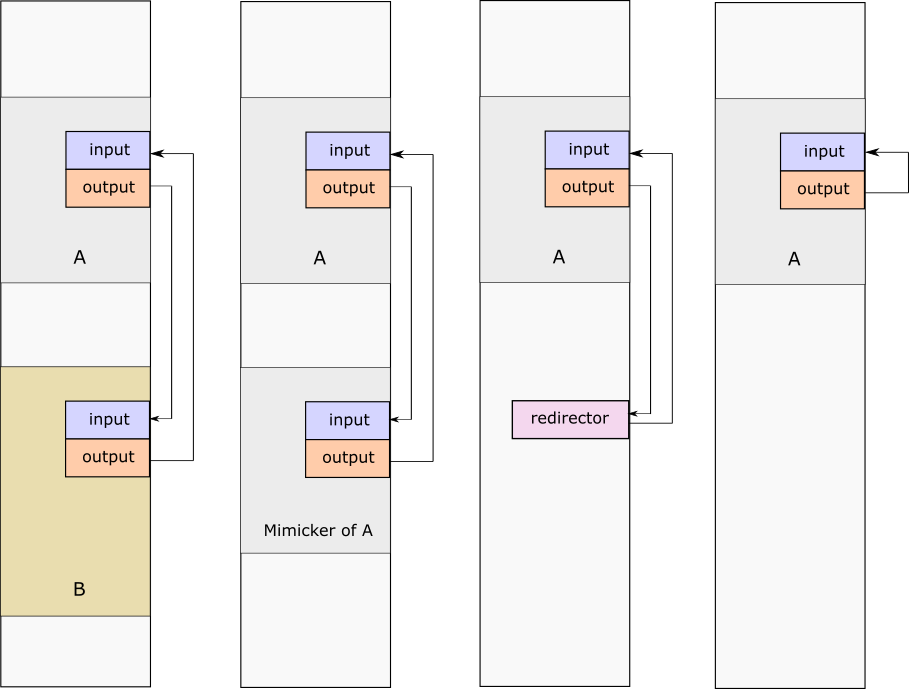}
            \caption{Conceptual generalization}
            \label{fig:text}
    \end{subfigure}
		\caption{Proposed methodology}
	\end{figure*}

The main contribution of this paper lies in the conceptual generalization of the conventional mirror test, namely going from Fig.~\ref{fig:vis} to Fig.~\ref{fig:text}. To be able to make meaningful analysis, a conversational agent replaces a robot in this context. Performed actions now correspond to output sent, and observing performed actions then correspond to the input received in the textual/conceptual version. According to this conceptual generalization, there is no restriction on the form of the agent as long as it has an input/output interface and an embodiment in any form. The need for an embodiment besides an input/output interface is discussed later on in the text.  

The proposed methodology consists of four stages as depicted in Fig.~\ref{fig:text}. For simplicity, a turn-based chatting session is assumed as usual for all stages. The first (leftmost) stage is the default (and conventionally the only) case considered in a chatting session. Chatbot A sends its output to chatbot B and receives B's output through its input. As conventionally the only case, chatbot A is aware of the fact that it is communicating with another entity (whether it be another chatbot or a human), and possibly has no motivation to think otherwise. Note that, in nearly all of the cases relating two distinct entities, there exists an intermediate communication channel, but is discarded here for simplicity. Then in simplistic terms, in Stage 1, A and B depict instances of two distinct programs with a corresponding input/output relation as necessary. 

As the first effective stage of the proposed methodology, the next stage replaces B with another instance of A, called the mimicker of A. Therefore, in this case there are two instances of the same program talking to each other. So the question then is: Will the agent A be able to recognize such a case and figure out (in an unsupervised manner) that the entity contacted is in fact an instance of itself instead of being an instance of a distinct program?

At this point, referring back to the visual version will be helpful. Since implementing such a stage for living beings is impossible in practice, a theoretical consideration is taken. Assume that there is an exact replica of the subject, and such replica can mimic any action taken in real-time without any delay. What will be the consequences when these two entities face each other? Interestingly, if the subject contains non-determinism, such mimicry will break at some point. Will such mimicry continue on till infinity otherwise? Not necessarily. Although these are instances created from the same source, they may be subject to change as interaction goes on and such change may differ for each. After all they are not exactly the same, at least they have different coordinates in the system they belong to (consider a system in which a change mechanism that depends on the coordinates of the individuals exists)

Going back to the conceptual version, if determinism and no-change policies hold, A will receive responses from its mimicker, exactly the responses it would give. Hypothetically, A can first send the query to itself (an ability to be formalized later in Stage 4) and then to the mimicker, and can check for the equality of those two responses. However, it seems as if such procedure should be repeated till infinity for a perfect mimicker to be detected and this dilemma paves way to the next stage. 

The third stage is the stage that conceptualizes the mirror test as depicted in the third column of Fig.~\ref{fig:text}, in which an input/output redirector is conceptually used as a mirror. In fact, from A's perspective, such stage initially seems indistinguishable from Stage 2. In visual version, a perfect mimicker behind a glass, and a reflection in a mirror rather sound as two observably equivalent scenarios. Then, how is it possible that a human or a capable animal is able to definitely grasp the concept of reflection without needing to test till infinity as Stage 2 demands? When and how does the necessary ``click'' happen? 

Equally applicable to both visual and textual, such ``click'' happens when the subject figures out that in fact there are not two separate entities, but these two entities actually refer to the same physical space, namely to the subject itself, rather it be the body or the address space respectively. This is rather equivalent to figuring out the principles of a mirror or an input/output redirection. Namely in textual version, it is equivalent to figuring out that the contacted entity is not an actual entity but a reference that rather redirects to subject's own input. Note that, how this realization takes place is rather a deeper issue that has partly been addressed in Sec.~\ref{sraaa} for the visual version. Through Stage 4, the final stage, these deeper issues are to be introduced for the textual version this time.

Note that, up to this stage, input/output relations are assumed to be established through enforcement instead of choice. As an example, an online conversational agent is hopelessly forced to chat with anyone that connects to its interface and has to respond somehow to each query that stranger sends. Similarly, in Stage 2, a mimicker is instantiated and the necessary input/output relations are established without question, and such procedure also holds for Stage 3. In that sense, Stage 4 is actually a symbolic stage in which the agent A is depicted to have redirected its output to its own input, without needing an additional redirector as in Stage 3. A deep question then is: Will an agent passing Stage 3 be able to configure itself as depicted in Stage 4? In other words, will a successful textual self-recognition lead to further self-awareness abilities such as the concept of inner-voice, or inner-speech? Possible implications of this and related issues are to be discussed next. 

\begin{table*}[t!]
\caption{Comparison of visual versus textual/conceptual self-recognition and its relation to levels of self-awareness in children as proposed in~\cite{rochat2003five}. }
\label{tab:table2}

\begin{tabular}{p{0.05\textwidth}p{0.05\textwidth}p{0.38\textwidth}p{0.38\textwidth}}
\toprule

\textbf{Stage} & \textbf{Level} & \textbf{Visual explanation} & \textbf{Textual/Conceptual explanation}                                                                             \\ \midrule
1   & L-0 & Unaware of any reflection or mirror  & One continues chatting as if contacted one is another entity.\\
2   & L-1                 & One realizes the mirror is able to reflect things  & One realizes there is copy/pasting going on.  \\
    & L-2                 & One can link the movements on the mirror to what is perceived within their own body & One realizes that copy/pasting is rather continuous.                                                                           \\
3   & L-3                 & One has the new ability to identify self. What is in the mirror is not another person but actually oneself & One realizes that one chats with oneself.          \\
4   & L-4                 & One can recognize oneself as a permanent self. & One can recognize that one is a program residing in an address space of a computer. \\
    & L-5                 & One understands that one can be in minds of others. & One can understand that one is an agent in a digital world. \\ \bottomrule
\end{tabular}
\end{table*}

\subsubsection{Implications of textual/conceptual self-recognition.}

The original visual version has partially been discussed in Sec.~\ref{sraaa}, but there is more to be mentioned. Passing the mirror test for the first time possibly grants the subject the ability to virtually place the observer (i.e. the virtual eye) in a $3^{rd}$ person manner, providing a form of perspective-taking. In other words, a mirror (theoretically directable towards anywhere) provides a different viewing perspective as if the subject is virtually somewhere else, or equivalently granting the subject to both observe and also transmit its actions to otherwise inaccessible portions of the world. Similarly, through an input/output redirector it is possible to bind the input/output of the subject theoretically anywhere in memory. 

It is still in question whether passing Stage 3 automatically grants the agent the ability to redirect its output as desired without needing any tool, whether it be to its own input or to somewhere else. The validity of such a possible implication should be carefully considered in a formal manner. However, assuming that the agent can redirect its output to its own input, whether it be through an internal mechanism or an external one as depicted in Stage 4, then this can serve as a formal geometrical definition of the inner-voice concept and can lead to further self-awareness discussion~\cite{chella2020developing}. The implications of having an inner-voice are truly far-reaching in return~\cite{morin2011self,alderson2018varieties}. 

In a related manner, if an agent is able to consider actions without being actually performed and how they would be observed even in the absence of a mirror, this rather translates to a form of imagination. In the most general setting, there happens to be the ability to imagine physically non-existing (and possibly dynamic) scenes from an arbitrary perspective. Then in the textual version, this would correspond to the ability of virtually creating non-existing entities and having hypothetical encounters/conversations with them. This can then possibly be tied to the well-entertained theory of mind concept, in which non-existent entities are then models, previously created by the subject, of actual entities. Note that, there still exists an ongoing debate on whether self-recognition and theory of mind concepts can be regarded as highly correlated or not~\cite{morin2011self}, but a gradual connection can definitely be made as apparent in our example.

\subsubsection{Relation to visual version and to levels of self-awareness.}

As noted before, self-recognition might be a gradual concept rather than being binary, and may be related to a developmental procedure. Such a model has been proposed for the development of self-awareness in human children~\cite{rochat2003five}. According to such a model there are 6 levels as depicted in Table.~\ref{tab:table2}. Corresponding stages in our methodology and their visual/textual/conceptual counterparts are also listed. Note that this is rather a preliminary sketch of possible relations between self-recognition and levels of self awareness.

\subsubsection{A comparative analysis.}
A comparative analysis is now given with respect to TT referring back to Sec.~\ref{sec:att}. It is not possible to fully address each item presented in that section, as a more formal and rigorous version of our methodology should be devised for such a detailed consideration. Addressing the first item, it is questionable whether consistent logical systems will be capable of self-recognition or not or whether paraconsistency is a requirement for self-recognition, but such related claims need to be considered rigorously as a future perspective. However, as a self-contained test devoid of external dependency, our methodology is capable of detecting not just human-specific but a general form of intelligence, perhaps being a precursor for higher forms. There is no external judge that specifies the outcome, but the testee judges oneself in a truly objective manner. An observer is in fact needed to record the success or failure of the test, but the outcome of the test is independent of whether an observer exists. As a final related note, there might be a reasonable time limit (or number of turns) for the testee to be deemed successful or not.  

\section{Discussion and Conclusion}
Although studies based on the visual version exist as mentioned previously, to our knowledge there has been no research on the textual version of the mirror test as proposed here. It is also a big question whether conversational agents, which have previously passed the Turing test, will also be able to fulfill all of the proposed stages depicted here. 

Two approaches come to attention as possible solutions to this textual version. A conventional solution can be to integrate concepts that mirror neurons provide with natural language processing tools, thus providing solutions parallel to the ones mentioned in~\ref{rsr}. As a mirror neuron is sensitive to a form of action, in textual version, this corresponds to a neuron sensitive to a pattern of text (whether it be an input or output). This way, an agent may able to detect that it is talking to oneself if it ``has a neuron'' that is sensitive to patterns of text that are specific to the agent. 

Another solution may be to figure out a way to query the address of the memory cells one occupies in an unsupervised way. That way, by adequate querying one can then differentiate whether the entity contacted is another physical entity or in fact oneself. This way a solution for even detecting a perfect mimicker is possible. However, as noted the agent needs to seek this address querying behavior oneself. If such address querying mechanism is hard-coded then it defies the goal of the test to begin with. It would then correspond to implanting a perfectly working mirror self-recognition chip into a monkey's brain and watch the monkey recognize oneself in the mirror.

\section*{Declarations}

\subsection*{Funding}
Not applicable
\subsection*{Conflict of interest/Competing interests}
The authors declare that they have no conflict of interest
\subsection*{Availability of data and materials}
Not applicable
\subsection*{Code availability}
The work is based on a conceptualization. No code is written in the process.
\subsection*{Authors' contributions}
Conceptualization: Yigit Oktar; Methodology: Yigit Oktar, Erdem Okur; Literature search: Yigit Oktar, Erdem Okur; Writing – original draft preparation: Yigit Oktar, Erdem Okur; Writing – review and editing: Erdem Okur, Mehmet Turkan; Supervision: Mehmet Turkan.
\subsection*{Ethics approval}
Not applicable
\subsection*{Consent to participate}
Not applicable
\subsection*{Consent for publication}
Not applicable






\bibliography{refs}


\end{document}